\documentclass[runningheads]{llncs}
\usepackage[T1]{fontenc}
\usepackage{perpage} 
\MakePerPage{footnote}

\usepackage{graphicx}

\usepackage{enumitem}  
\usepackage{comment}

\usepackage{marvosym}   

\newcommand{\ie}{\textit{i.e.} }
\newcommand{\eg}{\textit{e.g.} }

\newcommand{\hatpi}{{\hat\pi}}

\newcommand{\iset}{\{I_\hatpi\}}
\newcommand{\yset}{\{Y_\hatpi\}}
\newcommand{\Ex}{E_\mathcal{\!\:\!\scriptscriptstyle X\!\:\!}^{\:\!t}}
\newcommand{\En}{E_{\!\,\!\scriptscriptstyle \Delta\!\:\!\!\:\!}^{\:\!t}}
\newcommand{\Fy}{F_\mathcal{\!\:\!\scriptscriptstyle Y\!\:\!}^{\:\!t}}
\newcommand{\Fi}{F_{\!\:\!\scriptscriptstyle I\!\:\!}^{\:\!t}}

\usepackage{subcaption}

\usepackage{tikz}

\usepackage{algorithm}
\usepackage{algpseudocode}

\usepackage{hyperref}
\hypersetup{
    colorlinks=true,
    linkcolor=blue,
    filecolor=magenta,      
    pdftitle={360x},
    pdfpagemode=FullScreen,
    }

\usepackage{amsmath}
\usepackage{amsfonts}
\usepackage{mathrsfs}
\usepackage{wrapfig}
\usepackage{scrextend}
\usepackage[para, flushmargin, multiple, ragged]{footmisc}
\usepackage{graphicx}
\usepackage{lmodern}
\usepackage{booktabs}
\usepackage{amssymb}   
\usepackage{pifont}    
\usepackage{wrapfig}
\usepackage{makecell}
\usepackage{amsmath}
\usepackage{bm}
\usepackage{multirow}

\usepackage{colortbl}
\usepackage{xcolor}

\usepackage{booktabs}
\usepackage{threeparttable}
\usepackage{multirow}
\usepackage{bm}
\usepackage{nameref}
\usepackage{makecell}
\usepackage{enumerate}

\usepackage{tabularray}

\begin{document}

\title{ Domain Game: Disentangle Anatomical Feature for Single Domain Generalized Segmentation}

\author{Hao Chen \inst{1} \and 
        Hongrun Zhang \inst{1} \and U Wang Chan \inst{2} \and 
        Rui Yin \inst{3} \and \\ Xiaofei Wang\inst{1} \and  Chao Li \inst{1, \textrm{\Letter}}  }   %

{\let\thefootnote\relax\footnotetext{\textrm{\Letter}Corresponding author.} }

\institute{Department of Clinical Neurosciences, University of Cambridge, UK 
 \and
Department of Computer and Information Science, University of Macau, China \and
Department of Sports Medicine and Joint Surgery, Nanjing First Hospital, \\Nanjing Medical University, China
}

\maketitle       

\definecolor{airforceblue}{rgb}{0.36, 0.54, 0.66}
\definecolor{rpink}{RGB}{255,102,204} 

\newcommand{\hao}[1]{{\color{purple}Hao: #1}}

\begin{abstract}  



\textit{Single domain generalization} aims to address the challenge of out-of-distribution generalization problem with only one source domain available. Feature distanglement is a classic solution to this purpose, where the extracted task-related feature is presumed to be  resilient to domain shift. However, the absence of references from other domains in a single-domain scenario poses significant uncertainty in feature disentanglement (\textit{ill-posedness}). In this paper, we propose a new framework, named \textit{Domain Game}, to perform better feature distangling for medical image segmentation, based on the observation that diagnostic relevant features are more sensitive to geometric transformations, whilist domain-specific features probably will remain invariant to such operations. In domain game, a set of randomly transformed images derived from a singular source image is strategically encoded into two separate feature sets to represent diagnostic features and domain-specific features, respectively, and we apply forces to pull or repel them in the feature space, accordingly. Results from cross-site test domain evaluation showcase approximately an \textasciitilde11.8\%  performance boost in prostate segmentation and around \textasciitilde10.5\% in brain tumor segmentation compared to the second-best method. {\color{rpink} The codes will be available at \textit{GitHub}.}



\keywords{Model  generalisation  \and Feature disentanglement \and Deep segmentation.

}
\end{abstract}

\section{Introduction}

\noindent\textbf{Background:} 
\textit{Medical image segmentation} serves to classify anatomical structures \cite{hussain2022modern} within organs or lesions at the pixel level.  Despite documented success across various applications, significant challenges remain, where segmentation models may fail catastrophically when presented with out-of-distribution data. This challenge is commonly referred to as  \textit{domain shift} \cite{kondrateva2021domain}, stemming from variations of intensity distribution across datasets acquired from different protocols or procedures at multiple clinical centers \cite{guan2021domain}.

In this study, we examine a practical but challenging scenario in addressing \textit{domain shift}, attributed to its constrained availability of a single domain (source) for training. This scenario, commonly known as single-domain generalization (SDG) \cite{ouyang2022causality}, aims to improve generalization capabilities towards multiple novel domains (target). 
Contrary to the multi-domain approaches that can leverage the similarities/variations between different domains, SDG lacks the essential contrastive information to guide the learning process,  thus posing challenges in attaining robust generalization across domains. \\

\noindent\textbf{Related Work:} Studies concerning the SDG  challenge can be generally categorized into two groups that focus on either data or features. Data-based approaches aim to bridge diverse domains or enhance training diversity through data augmentation or generation \cite{ouyang2022causality,volpi2018generalizing,xu2022adversarial}. For instance, \textit{Ouyang et al.} \cite{ouyang2022causality} utilize a randomized network functioning as a style generator to develop artificial domains tailored specifically for training from a single source. Despite being effective in managing minor distribution changes, data-based approaches are vulnerable to substantial domain shifts beyond their training strategy.

Feature-based methods are designed to utilize domain-invariant features to enhance model resilience against unforeseen out-of-domain influences \cite{sun2021recovering,dou2019domain}. Central to feature-based approaches is the feasibility of distinguishing domain-invariant features ($X$) from domain-specific features ($\Delta$) within an intricate high-dimensional feature space ($E(I)$) derived from a given image $I$:
\begin{equation}
    E(I) = X \oplus (\Delta), 
    \label{eq:illpose}
\end{equation}

\noindent where $E(\cdot)$ signifies a mapping function that transforms an image into its representation within feature space and $\oplus$ denotes the feature fusion operation. 


It is of note that surrounding $\Delta$ with parentheses indicates that in some methodologies \cite{zhao2020domain,zhao2021robust,DSU}, the domain-specific features are not explicitly delineated. For instance, the \textit{feature alignment} approach \cite{zhao2020domain,zhao2021robust}serves as an efficient strategy for aligning the distributions of $X$ across source and target domains, thereby mitigating the domain discrepancies.  Similarly, the \textit{domain shifts with uncertainty} (DSU)  \cite{DSU} integrates feature statistics for modeling the uncertainty of $X$ through analyzing multiple augmented examples, aiming to alleviate the model uncertainty on the target domain.  However, these methods are challenged by the ill-posed nature\footnote[1]{This denotes the existence of more than one possible solution to the equation.}  emphasized by Eq. \textcolor{red}{\ref{eq:illpose}}, rendering isolating only $X$ insufficient for ensuring its consistency across domains.


In response to the ill-posed challenge, \textit{feature disentanglement} offers an alternative approach that explicitly models $X$ and $\Delta$ and then distinguishes them \cite{chen2023multi,Mao2023}. For example,  a recent endeavor \cite{BayeSeg} proposes that $\Delta$ conforms to a normal distribution, wherein $\Delta \in \mathcal{N}(\mu, \text{diag}(\sigma)^{-1})$, with both $\mu$ and $\sigma$ being trainable parameters. By holding $\Delta$ constant, Eq. \textcolor{red}{\ref{eq:illpose}} turns deterministic. However, the Gaussian prior assumption imposes strict constraints on the underlying distribution, which
may not effectively capture the complex and varied nature of the \textit{domain shift}, potentially leading to oversimplified representations \cite{landry2019cello}. It is worthnoted that \textit{Gu et al.} \cite{gu2023cddsa} propose a disentangle framework that utilize style augmentation to generates image with same pathology structure but in different style, \ie images with same $X$ but with different $\Delta$ to learn to seperate domain-agnostic $X$. While we observe that the domain shift may not just the intensity distribution that relates to the machine, but also has population-specific shift that may affect the distribution of the atonomic feature, \eg patients from different continent or in different age groups. Drawing from this attention, we propose in this paper to learn the diagnostic feature in the highlight learning process on the by interference of $X$ to learn to extract the $X$ from images samples have different $X$ but share same domain $\Delta$, thus more focus on the atonomy structure distribution. \\

\noindent\textbf{Motivation and Contributions:} Following the aforementioned analysis, we seek to develop a feature disentanglement method that does not rely on explicit assumptions about the underlying distributions of either $X$ or $\Delta$. Clinically, the \textit{diagnostic features} of organs or lesions (\eg mass tumors)  are consistently associated with local anatomical structures across domains, \ie domain-invariant features. However, we observe that these structural intricacies in a specific image are responsive to geometric transformations; \eg rotating an image will also rotate its geometric information. Conversely, the \textit{domain-specific features} are more associated with the overall intensity distribution. Given that transformations do not alter this distribution, domain-specific features remain unaffected by geometric transformations.

\begin{figure}[t]
    \centering
    \includegraphics[width=1\linewidth, trim=0 455 700 4, clip]{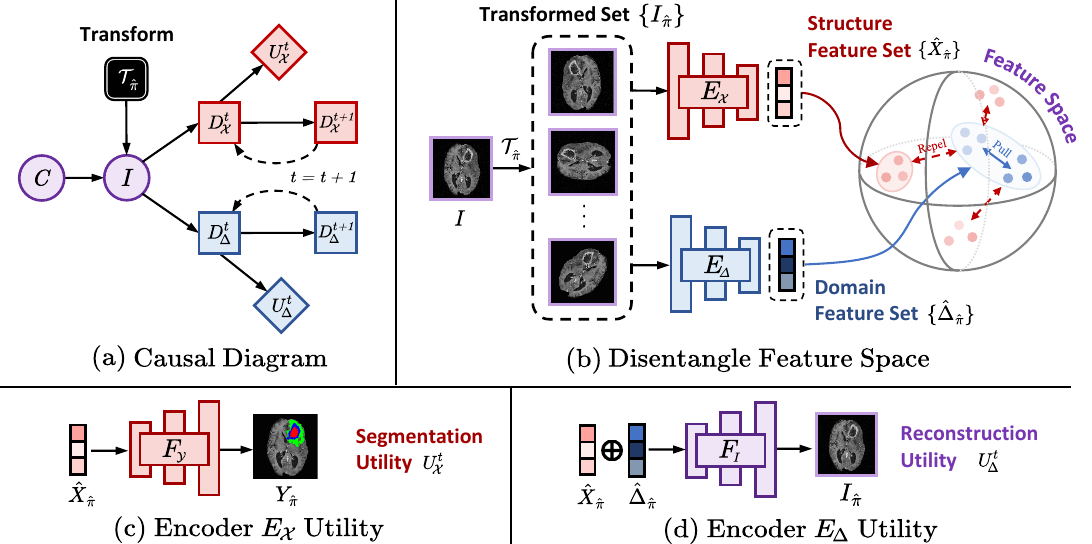}
    \caption{ \textbf{Domain Game overview.} 
    Colours indicates the relation to distinct features, \ie  purple - overall, red - diagnostic and blue - domain-specific features.
    (a) Image  $I$ is drawn from singular domain ($C$) to update the parameters $D$ at each step according to utilities $U$. 
    (b) Transformed set is generated by applying $\mathcal T_\hatpi$ to $I$,  and subsequently encoded into feature sets by two encoders. Then features within $\{\hat \Delta_\hatpi \}$ are pull together to learn domain-specific character and introduce repulsion between $\{\hat X_\hatpi \}$ and $\{\hat \Delta_\hatpi\}$ to enforce disentanglement. (c) The quality of $\hat X_\hatpi$ is measeured through segmentation utility. 
    (d) The disentanglement is achieved by fusing randomly paired $\hat X$ and $\hat \Delta$ to reconstruct the input. }
    \vspace{-0.6cm}
    \label{fig:arch}
\end{figure}

In this paper, we introduce a novel feature disentanglement framework within a single domain.  We elucidate this framework using an analogy to combinatorial games to leverage the distinct feature characteristics, thereby introducing our concept called the \textit{Domain Game}. The contributions of this work are threefold:

\begin{itemize}[topsep=1mm, left=8pt, itemsep=1mm] 

  \item We propose \textit{Domain Game}, a new paradigm in single-domain generalization that seeks to address the inherent ill-posed challenge of feature disentanglement by leveraging our prior knowledge of diagnostic feature characteristics.
  
  \item To emphasize the distinct feature properties for facilitating disentanglement, the \textit{Domain Game} introduces these properties as update utilities specific to different networks. Intriguingly, two networks strategically intertwine to disentangle the feature space,  reflecting an analogy with combinatorial games.

  \item We perform extensive experiments and validate that the proposed method effectively produces more robust features and outperforms state-of-the-art models on out-of-distribution test domains.
\end{itemize}

\section{Methodology}

Figure \ref{fig:arch}{\color{blue}a} illustrates the overview causal diagram of the \textit{Domain Game}. The context variables ($C$) symbolize the single domain where the image ($I$) is sampled at each iteration step. As indicated in Figure \ref{fig:arch}{\color{blue}b}, \textit{Domain Game} introduces two encoders  ($\Ex$, $\En$) to learn the disentangled feature sets. The parameters ($D$, squares) and utility variables ($U$, diamonds) associated with these encoders are indicated by the same color. For each iteration, both encoders compute their utility functions $U^t$ based on the current parameter $D^t$ and evolve to an optimized parameter $D^{t+1}$ in the next iteration ($t+1$). The algorithmic flow of Domain-Game is illustrated in Algorithm \ref{alg:game}, and we will introduce its key steps in the following sections.\\

\begin{algorithm}[t]
\caption{Domain Game}\label{alg:game}
\begin{algorithmic}
\Require paired set $\{I, Y\}_\text{set}$, iterations $N$, transform $\mathcal{T}_{\hat\pi}$

\State initialize encoders $\Ex$ and $\En$ and decoder functions $\Fy$ and $\Fi$
\While{$N \neq 0$}
\State sample $I, Y\in\{I, Y\}$
\State apply  $\iset\gets\mathcal{T}_{\hatpi}(I)$, $\yset\gets\mathcal{T}_{\hatpi}(Y)$ \Comment{geometric transformation}

\State get features iteratively, loop every $I_\hatpi\in\iset$

\hspace{0.9cm}$\hat X_\hatpi\gets\Ex(I_\hatpi)$, $\hat \Delta_\hatpi\gets\En(I_\hatpi)$   



\State calculate  $\mathcal{L}_{\scriptscriptstyle \mathcal{FD}}$  \Comment{\textit{Step 1} feature disentanglement}

\State \textbf{Updates Encoder $\Ex$ parameters}:
  \State  \hspace{1cm}   get $\hat Y_\hatpi\gets\Fy(\hat X_\hatpi)$
  \State  \hspace{1cm}   calculate $U_\mathcal{\!\:\!\!\:\!\scriptscriptstyle X\!\:\!}^t \gets \text{Dice} (Y_\hatpi, \hat Y_\hatpi)$  \Comment{\textit{Step 2}  segmentation utility}
    
\State \textbf{Updates Encoder $\En$ parameters}:
    \State \hspace{1cm}  get $\hat I_\hatpi\gets\Fi(\hat \Delta_s | \hat X_\hatpi)$, $\Delta_s$ sampled from $\{\hat\Delta_\hatpi\}$
    \State \hspace{1cm}  calculate $U_{\!\,\!\!\scriptscriptstyle \Delta\!\:\!\!\:\!}^t\gets \text{PSNR}(I_\hatpi, \hat  I_\hatpi)$  \Comment{\textit{Step 2} reconstruction utility}


\State $N \gets N - 1$
\EndWhile
\end{algorithmic}
\end{algorithm}



 
\noindent \textbf{\textit{Step 1.} Feature Disentanglement.}
Conventional feature learning typically focuses on establishing a direct one-to-one mapping between images and corresponding features, denoted as $I \to X$. While this method is straightforward, the amalgamation of diverse image aspects into a singular feature frequently poses challenges for models to effectively operate across varied domains. 

To address this constraint, we propose to separate the anatomical features crucial for the diagnostic process from the domain features inherent in the captured image properties. Our clinical observations reveal that the geometry description is prominent in describing anatomical structures, \eg shape, size, orientation, and spatial relationships. Thus, the local anatomical features are responsive to geometric alterations, \ie geometry transformations will alter the structures, resulting in distinctive features. On the other hand, the image property (ref as domain feature in medical images) maintains its integrity when subjected to geometric transformations.

We leverage the the different responsive level of geometry transformations of anatomical feature $X$ and domain feature $\Delta$ to perform disentanglement. In essence, we cultivate a variable $\hat{X}$ that reveals $\mathcal{T}$-sensitive configurations to retain anatomical structure information. Simultaneously, we identify $\mathcal{T}$-insensitive traits for segregating domain features into $\hat{\Delta}$.


Specifically, we employ a multitude of viewpoints through diverse randomly transformed instances to offer a unique estimation for features that exhibit within ill-posed nature. At each interation, a set of $n$ geometric transformations denoted as $\{\mathcal{T}_{\hatpi_i}\}$ is  stochastically sampled, where $\hat{\pi}$ signifies the transformed effect, and $i \in \mathbb{N}: 0 < i \leq n$ represents the index.  Subsequently, every transformation $\mathcal{T}_{\hat{\pi}_i}$ is applied to the image-label pair $(I, Y)$, generating the transformed pair $(I_{\hat{\pi}_i}, Y_{\hat{\pi}_i})$. The encoders then derive $\hat{X}_{\hat{\pi}_i}$ and $\hat{\Delta}_{\hat{\pi}_i}$ from the input $I_{\hat{\pi}_i}$ based on current parameters set $D^t$, as illustrated in Figure \ref{fig:arch}{\color{blue}a}:


\begin{equation} 
\mathcal{L}_{\scriptscriptstyle \mathcal{FD}} = \sum_{i}^n { \left\| \mathcal{T}_{\hatpi_i}(\hat{X})  - \hat{X}_{\hat{\pi}_i} \right\|^2} + \sum_{i}^n \sum_{j\neq i}^n\left\| \hat{\Delta}_{\hat{\pi}_j}  - \hat{\Delta}_{\hat{\pi}_i} \right\|^2.
 \label{eq::dis}
\end{equation}

Within this framework, the symbol $\hat{X}$ signifies the anatomical feature from the unaltered input $I$. To ensure consistent alignment between anatomical structures, it undergoes a congruent transformation denoted as ${\mathcal{T}}_{\hat{\pi}}$, in accordance with  $I_{\hat{\pi}_i}$. The transformation ${\mathcal{T}}_{\hat{\pi}}$ acts on the spatial dimensions represented by $\mathbb{R}^{H\!\times\!W}$ within the input $\hat{X}$, which is structured as $\mathbb{R}^{C\!\times\!H\!\times\!W}$.

\noindent \textbf{\textit{Step 2.} {Posterior Utility Maximisation.}}
The diagnostic features $\hat X_{\hat{\pi}}$ produced by the encoder $\Ex$ are utilized by the decoder function $\Fy(\cdot)$ for segmentation purposes. The performance of this segmentation process is assessed using the Dice coefficient, which quantifies the utility as:

\begin{equation}
U_\mathcal{\!\:\!\!\:\!\scriptscriptstyle X\!\:\!}^t = \text{Dice} (\Fy(\hat X_\hatpi), Y_\hatpi).
\end{equation}

Encoder $\En$ aims to isolate domain features that are not exclusively tied to a particular image. This is achieved by randomly sampling a $\hat{\Delta}_s \in \{\hat{\Delta}_{\hat{\pi}}\}$ and pairing it with a $\hat{X}_{\hat{\pi}}$ to 
 reconstruct the $I_{\hatpi}$ using the decoder $\Fi(\cdot)$. 
Given that $\hat{X}_\hatpi$ is associated with a specific $I_{\hatpi}$ while $\hat{\Delta}_s$ is not, $\hat{X}_\hatpi$ serves as the condition for specifying the reconstruction target. The utility of $\En$  can be assessed by the reconstruction fidelity using the Peak Signal-to-Noise Ratio (PSNR):

\begin{equation}
U_{\!\,\!\!\scriptscriptstyle \Delta\!\:\!\!\:\!}^t=\text{PSNR} (I_\hatpi, \Fi( \hat\Delta_s  | \hat X_\hatpi)),\ \hat\Delta_s\in \{\hat\Delta_\hatpi\}.
\end{equation}

Additional information and implementing details regarding the detailed optimization process can be found in the Supplementary.


\noindent \textbf{\textit{Analogy} to Combinatorial Game.}  
Utility maximization and loss minimization are closely related concepts; for example, maximizing the Dice can be seen as minimizing 1 - Dice. While this is true, we want to emphasize that utility represents the level of satisfaction, with higher being better.

In the game of Go\footnote{A  classic example of a combinatorial game.}, two players aim to secure more territory than their opponent through adversarial maneuvering to displace the opponent and define boundaries, strategically dividing the board into distinct subspaces with established borders.  Drawing an analogy to this concept, the feature space of the \textit{Domain Game} can be viewed as being analogous to a game board. In this context, two players (represented by networks) endeavor to partition the board into two subspaces representing $X$ and $\Delta$, respectively. The feature space constraints (implemented by the Lasso penalty) function as the rules that confine the players to the board, while the repel force functions as the strategic moves that aim to displace the opponent to seize additional territory.

\subsection{{Implementing Details}}
This segment delineates the network architecture specifics and training approach for the \textit{Domain Game}. Two encoders are represented by EfficientNet-B2 \cite{tan2019efficientnet}, following the setting of BayeSeg \cite{BayeSeg}. The decoder functions employ DeeplabV3 \cite{chen2018encoder} for segmentation or reconstruction inference. We utilize a sliding window input with a window size of 3 along with stride of 1. The label of this window is aligning the target to the center of the window.  The optimization utilizes the AdamW optimizer with a learning rate set to 1e-4 for 1200 epochs. Furthermore, the learning rate decay mechanism adheres to a CosineAnnealing \cite{loshchilov2016sgdr} schedule with a time span of 30 units.

The absence of constraints within the feature space frequently leads to its scattered and sparse attributes. Recent study \cite{zhu2021geometric} indicates that the presence of an unknown domain shift can cause features to reside in the areas that lack meaningful interpretation within the learned feature space, consequently introducing a higher level of uncertainty. To mitigate this, we employ a lasso penalty on the feature space to eliminate redundant dimensions, thereby imposing shrinkage on the feature space effectively.

\begin{equation}
 \mathcal{L}_\text{Lasso} =\|\hat X + \hat\Delta\|, 
\end{equation}

\noindent where $\hat X = \Ex(I)$ and $\hat \Delta = \En(I)$ are derived based on current parameters $D_\mathcal{\!\:\!\!\scriptscriptstyle X\!\:\!}^t$ and $D_{\!\,\!\!\scriptscriptstyle \Delta\!\:\!\!\:\!}^t$ , correspondingly, at iteration $t$.\\

The optimization process is conducted across the entirety of the \textit{Domain-Game}, utilizing a \textit{minmax} strategy to minimize the loss associated with constraining the features, while simultaneously maximizing the utility functions for segmentation or reconstruction. This optimization formula can be  expressed as:

\newcommand{\Ux}[0]{U_\mathcal{\!\:\!\!\:\!\scriptscriptstyle X\!\:\!}^t}
\newcommand{\Un}[0]{U_{\!\,\!\!\scriptscriptstyle \Delta\!\:\!\!\:\!}^t}

\begin{equation}
   \underset{D_\mathcal{\!\:\!\!\scriptscriptstyle X\!\:\!}^t,  D_{\!\,\!\!\scriptscriptstyle \Delta\!\:\!\!\:\!}^t,
   P_{\!\:\!\scriptscriptstyle \mathcal{Y}}^t, 
   P_{\!\:\!\scriptscriptstyle I}^t}{\arg}  
   \min_{\mathcal{L}}  \max_{U}\ (\lambda\mathcal{L}_\text{Lasso} +  \mathcal{L}_\text{Pull} +   \mathcal{L}_\text{Repel} ) -
     (\Ux+ \omega\Un),
\end{equation}

\noindent where the lasso weight, denoted as $\lambda$, is determined as $\lambda = 5$ through experimentations to balance between feature redundancy and an overly simplified space. The norm weight of $\Un$ is denoted as $\omega$. This is introduced because the functions used in $\Ux$ (Dice) and $\Un$ (PSNR) have different value ranges. The weight is set as $\omega=5e^{-2}$ through experimentations. $P_{\!\:\!\scriptscriptstyle \mathcal{Y}}^t$ and $P_{\!\:\!\scriptscriptstyle I}^t$ are the parameter sets of the decoder functions $\Fy$ and $\Fi$, respectively.

\section{Experiments}
\subsection{Experimental Setup}


We evaluate our proposed method on two different tasks, including 1) the Prostate segmentation task and 2) the Brain tumor segmentation task. We compare four state-of-the-art approaches for domain generalization, \ie Cutout \cite{cutout}, IBN-Net \cite{ibn-net}, RandConv \cite{randconv}, DSU \cite{DSU}, and BayeSeg \cite{BayeSeg}. EfficientNet-B2 \cite{tan2019efficientnet} is used as the backbone for all methods and train for 1200 epochs for a fair comparison. In each source dataset, the data is randomly divided into training (70\%), validation (10\%), and testing (20\%). The model achieving the best performance on the source validation set is chosen for the target domain evaluation.

\vspace{0.2cm}
\noindent \textbf{Note:} The quantitative metrics for segmentation include the Dice coefficient and Jaccard similarity scores, reported on a scale of 10$e$-2. 
\vspace{-0.2cm}

\subsection{Prostate Segmentation}
Prostate segmentation benefits accurate volume measurement and boundary estimation, which aids diagnostic procedures for prostate diseases \cite{Prostate3}. We employ three public datasets of T2-weighted MR images: NCI-ISBI 2013 \cite{Prostate1} (n=60), I2CVB \cite{Prostate2} (n=19), and PROMISE12 \cite{Prostate3} (n=37). The patient number of each dataset is denoted by $n$. The datasets are split into six domains depending on the medical centres from which they originate (RUNMC, BMC, I2CVB, UCL, BIDMC, and HK). We adhere to the setting in BayeSeg \cite{BayeSeg} for training and evaluating generalization on prostate segmentation task.


\vspace{1mm}
\noindent\textbf{Analysis:} In Table \ref{tab:prostate},  RandConv \cite{randconv} attains the highest dice in the source domain, while Domain Game reaches the smallest average performance drop at around \textasciitilde\textit{5.4}\% in the target domains. Notably, the BIDMC dataset exhibits a significant performance decline compared to other target datasets,  possibly impacted by its bias fields. In this context, our approach surpasses the second-best performance achieved by BayeSeg by approximately \textasciitilde\textit{11.8}\%.

\begin{figure*}[!t]

\newcommand\Ma{\includegraphics[width=0.71 \textwidth, trim=8 400 388 0,clip]}  
\newcommand\Mb{\includegraphics[width=0.31 \textwidth, trim=8 400 850 8,clip]}

	\begin{center}
 \resizebox{1.0\columnwidth}{!}{
		\begin{tabular}{c@{\extracolsep{0.2em}}c}

         \Ma{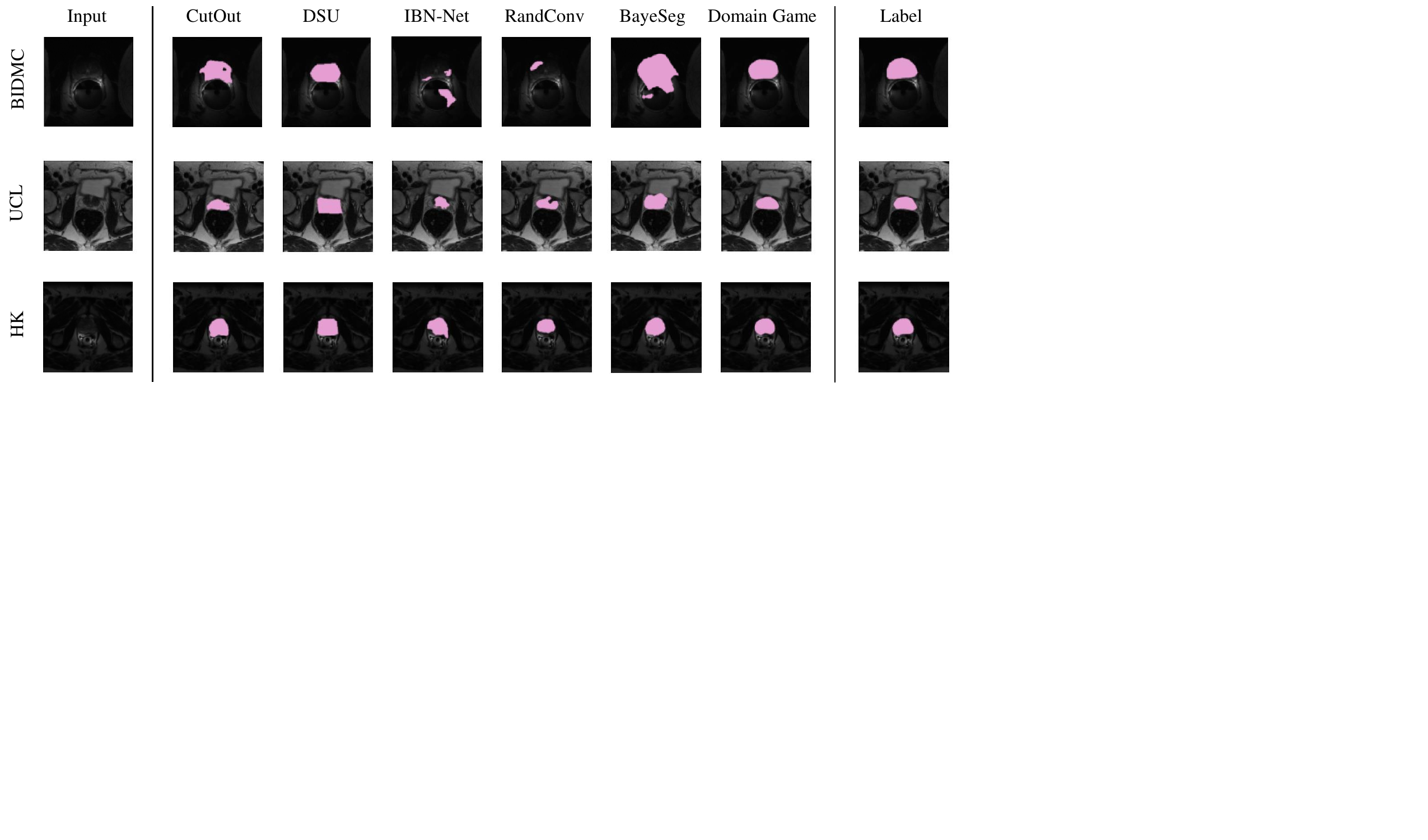}~ \qquad  &\Mb{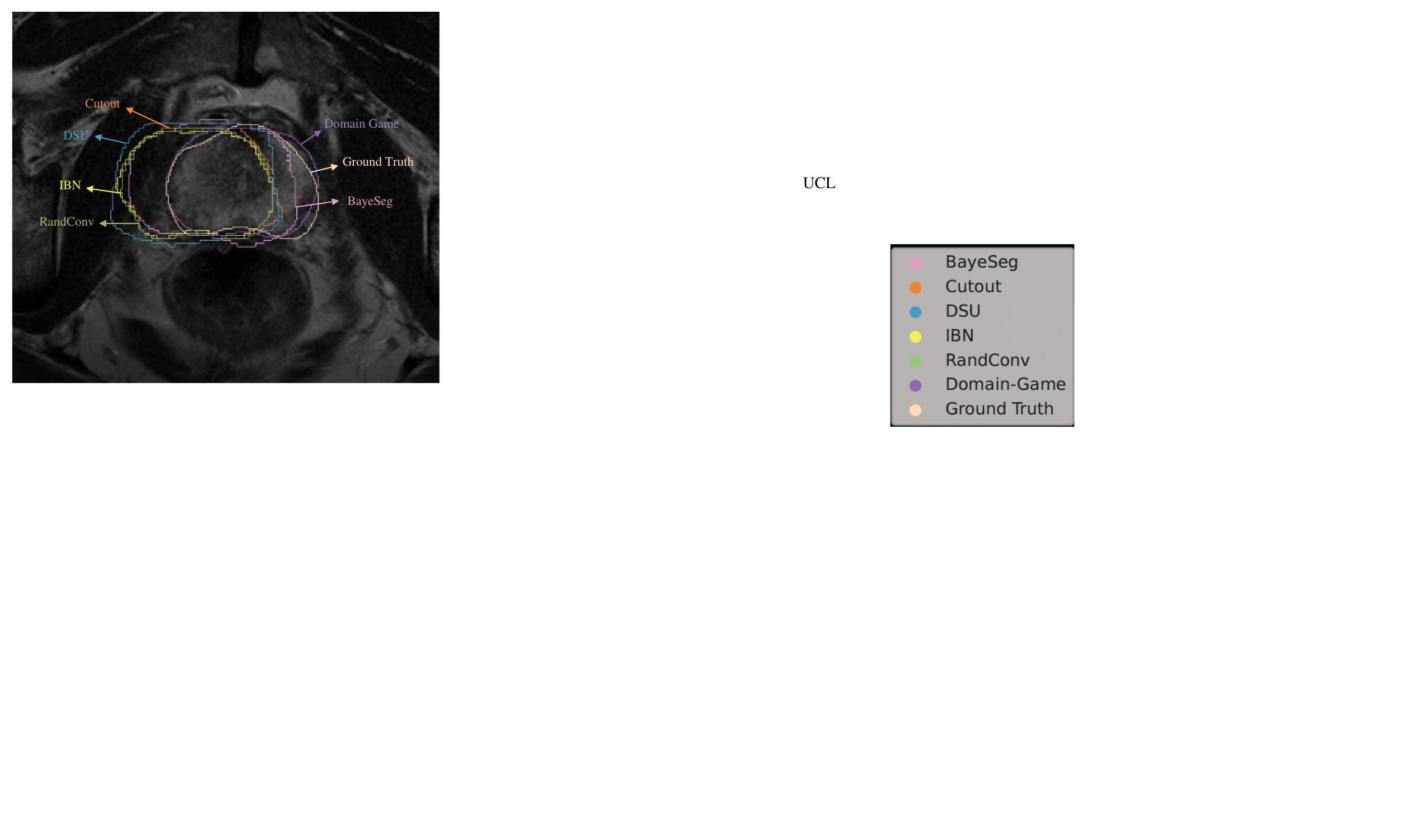} \\
         \vspace{-0.3cm}
 		(a) Visualization for cross-site prostate segmentation. ~ \qquad   & (b) Detailed comparison. \\
 	\end{tabular}
  }

	\end{center}
 
 \vspace{-0.3cm}
\caption{ \small Qualitative visualization through (a) broad and (b) intricate comparisons.}
  \vspace{-0.7cm}
\label{fig:prostate}
\end{figure*}

Fig. \ref{fig:prostate}{\color{blue}a} presents the qualitative segmentation results. Notably, DSU and Domain Game consistently deliver satisfactory outcomes, with Domain Game more resembling the labels.  Fig. \ref{fig:prostate}{\color{blue}b} presents a comprehensive comparative analysis within an example of the UCL domain. The results reveal that all the methods other than 
 Domain Game exhibit a leftward bias, while Domain Game notably preserves a closer alignment to the ground-truth label.



\newcolumntype{M}[1]{>{\centering\arraybackslash}m{#1}}
\newcommand{\Hone}{\hspace{0.2cm}}
\newcommand{\Hthree}{\hspace{0.4cm}}

\newcommand{\textBF}[1]{%
    \pdfliteral direct {2 Tr 0.3 w} 
     #1%
    \pdfliteral direct {0 Tr 0 w}%
}

\newcommand{\down}[1]{\textcolor[RGB]{ 180,12,22  }{\small(↓#1)}}

\begin{table*}[t]
\caption{Prostate segmentation performance across diverse target sites. The Dice score is reported in mean$\pm$std. The red down-arrow number indicates declined performance from source to target and lower values shows better generalization.} 
\label{tab:prostate}

 \centering
	\resizebox{\linewidth}{!}{
    \begin{tabular}{l@{\hspace{0.3cm}}|@{\hspace{0.3cm}}c@{\hspace{0.3cm}}|@{\hspace{0.3cm}}c@{\hspace{0.4cm}}c@{\hspace{0.4cm}}c@{\hspace{0.4cm}}c@{\hspace{0.4cm}}c@{\hspace{0.3cm}}|@{\hspace{0.3cm}}c}
     \toprule

    \multirow{2}{*}{Method}&  (Source) &
    &&\hspace{-0.7cm}\makecell{ (Cross-site Target) }\hspace{-0.7cm} &&
    & \multirow{2}{*}{\makecell[c]{Avg. on\\Target}} \\ 
    &RUNMC \cite{Prostate1} & BMC \cite{Prostate1}  &  BIDMC \cite{Prostate3} &  HK \cite{Prostate3}&  UCL \cite{Prostate3} &  I2CVB \cite{Prostate2} & \\
      \midrule

      Cutout \cite{cutout} & 85.88$\pm$04.2 & 79.42$\pm$06.9& 	54.42$\pm$13.3& 	77.17$\pm$06.8& 	79.75$\pm$03.2& 	78.99$\pm$06.2& 	73.95 \down{11.9}\\
        IBN-Net \cite{ibn-net} &  86.13$\pm$05.5	&  78.04$\pm$07.1	& 43.41$\pm$20.1	& 75.81$\pm$06.0	& 79.39$\pm$04.1& 	79.44$\pm$06.2	&  71.22 \down{14.9} \\
        RandConv \cite{randconv} &  \hspace{-0.1cm}\textBF{87.68$\pm$04.4} & 	83.75$\pm$05.0& 	54.35$\pm$15.8& 	77.89$\pm$06.4& 	83.76$\pm$05.1& 	77.21$\pm$12.3& 	75.39 \down{12.3} \\
        DSU \cite{DSU} &    85.49$\pm$07.2& 	77.89$\pm$06.1& 	66.14$\pm$12.5& 	75.77$\pm$07.4& 	78.49$\pm$04.1& 	75.89$\pm$12.8& 	74.84 \down{10.7} \\
        BayeSeg \cite{BayeSeg} &   86.57$\pm$05.7 & 		81.82$\pm$06.9	& 	 66.56$\pm$17.8	& 	 80.62$\pm$05.3	& 	 81.25$\pm$06.6	& 	 77.52$\pm$17.1	& 	 77.52 \down{09.1} \\
        \midrule
        Domain Game  &  86.92$\pm$05.3&	\hspace{-0.1cm}\textBF{85.42$\pm$05.4}	& \hspace{-0.1cm}\textBF{75.79$\pm$13.1} & \hspace{-0.1cm}\textBF{84.33$\pm$05.5}	& \hspace{-0.1cm}\textBF{86.10 $\pm$04.5 }& 	\hspace{-0.1cm}\textBF{79.50$\pm$05.2 }& 	\hspace{-0.1cm}\textBF{82.23 \down{04.7}}  \\

     \bottomrule
    \end{tabular}
    }
\end{table*}

\vspace{-0.3cm}
\subsection{Brain Tumor Segmentation}
Glioma, the most common malignant brain tumors, is known for high aggressiveness \cite{Li2023,Li2019}.  We aim to validate our method through brain tumor segmentation and investigate its utility across diverse geographic regions and age groups. 

We choose three datasets in this task, with Brats-Glioma \cite{baid2021rsna} ($n$=1251) as source domain,  Brats-Africa  \cite{adewole2023brain}  ($n$=60)  and BraTS-PEDs \cite{kazerooni2023brain} ($n$=228)  as target domains. Data from Brats-Glioma \cite{baid2021rsna}  patients were collected using various protocols and scanners across multiple institutions. The Brats-Africa  \cite{adewole2023brain} comprises data with unique characteristics, with low image contrast and resolution due to lower-quality MRI technology in areas surrounding Sub-Saharan Africa. BraTS-Pediatrics \cite{kazerooni2023brain} assembles pediatric brain tumor patients characterized by differing imaging and clinical profiles compared to adult brain tumors.

The brain tumor task segmentation glioma sub-regions: all portions of the necrotic tumor core (NTC $-$ label 1),  and the enhancing tumor (ET $-$ label 2). Although the BraTS 2023 dataset encompasses multi-parametric MRI sequences, we specifically select the post-gadolinium T1-weighted (T1Gd) sequence as it is frequently utilized in clinical practice for delineating tumor morphology.  We select enhancing tumors (ET - label 2) and the non-enhancing tumor core (NTC - label 1) as labels in our experiments. These two labels are annotated in both the Brats-Glioma and Brats-Africa datasets. However, Brats-Pediatrics  exclusively includes ET, and consequently, only ET is assessed in the Brats-Pediatrics  dataset \cite{kazerooni2023brain}. The BraTS 2023 dataset also includes region labels for edematous/invaded tissue. However, due to the challenges in clinically identifying this in T1Gd scans, we have opted to exclude this label from our experiments.






\begin{table*}[!tbp]
\caption{Brain tumor segmentation results reported in Dice  and Jaccard  scores. 
} %
\label{tab:brain}

 \centering
	\resizebox{\linewidth}{!}{

    \begin{tabular}{l@{\hspace{0.3cm}}|cc|cc|cc|cc}
    %

     \toprule
     
    \multirow{3}{*}{Method}&\multicolumn{2}{c|}{ (Source)}  
    &\multicolumn{2}{c|}{(Cross-site Target)}  
    &\multicolumn{2}{c|}{(Cross-age Target)}  
    &\multicolumn{2}{c}{\multirow{2}{*}{\makecell[c]{Avg. on\\Target}}}\\ 

    
    & \multicolumn{2}{c|}{BraTS-Glioma \cite{baid2021rsna}} & \multicolumn{2}{c|}{BraTS-Africa  \cite{adewole2023brain}}& \multicolumn{2}{c|}{BraTS-Pediatrics \cite{kazerooni2023brain}} & \\
        & Dice & Jaccard & Dice & Jaccard & Dice & Jaccard & Dice & Jaccard \\
    
      \midrule 
      Cutout \cite{cutout} 
      & 75.78$\pm$14.2 & 72.19$\pm$25.3 
      & 57.37$\pm$27.0  & 54.82$\pm$23.8	
      & 45.57$\pm$22.2 & 39.16$\pm$18.2
      & 51.47 \down{24.31} & 46.99 \down{25.20}\\
      
        IBN-Net \cite{ibn-net} 
        & 77.94$\pm$17.3 & 73.92$\pm$27.2
        & 62.31$\pm$26.5	&  51.75$\pm$20.6 
        & 54.46$\pm$22.4 & 42.03$\pm$20.3 
        & 58.39 \down{19.55} & 46.89 \down{27.03}\\
        
        RandConv \cite{randconv} 
        & \hspace{-0.1cm}\textBF{81.38$\pm$12.2} &\hspace{-0.1cm}\textBF{76.98$\pm$25.2}
        & 63.35$\pm$29.1 &	53.18$\pm$23.4 
        & 51.83$\pm$24.9	& 50.69$\pm$19.7
        & 57.59 \down{23.79} & 51.94 \down{25.05}\\
        
        DSU \cite{DSU} 
        & 79.23$\pm$14.6 & 75.31$\pm$26.4  
        & 61.09$\pm$28.7 & 52.62$\pm$24.9 
        &  55.12$\pm$21.5 &  52.78$\pm$19.6	  
        & 58.11 \down{21.12} & 52.70 \down{22.61} \\
        
        BayeSeg \cite{BayeSeg} 
        & 78.28$\pm$17.6  & 74.83$\pm$28.5 
        & 62.88$\pm$27.5	 &56.83$\pm$22.5	 
        & 51.89$\pm$21.7 & 51.65$\pm$16.1  
        & 57.39 \down{20.89} & 54.24 \down{20.59} \\
    
        \midrule
        Domain-Game   
        &\ \  78.47$\pm$17.3    \ \ &\ \ 	 75.49$\pm$29.1\ \ 
        & \ \ \hspace{-0.1cm}\textBF{69.73$\pm$26.3}    \ \ &\ \     \hspace{-0.1cm}\textBF{62.05$\pm$28.5} \ \ 
        &	\ \  \hspace{-0.1cm}\textBF{60.36$\pm$22.6}   \ \ &\ \ 	\hspace{-0.1cm}\textBF{57.02$\pm$17.2} \ \ 
        &	\ \ \hspace{-0.1cm}\textBF{65.05 \down{13.42}}  \ \ &\ \hspace{-0.2cm}\textBF{59.53 \down{15.95}} \\
     \bottomrule
    \end{tabular}
    }

\vspace{-0.4cm}
\end{table*}



\begin{figure}[t]
    \centering
    \includegraphics[width=1\linewidth, trim=8 285 388 0, clip]{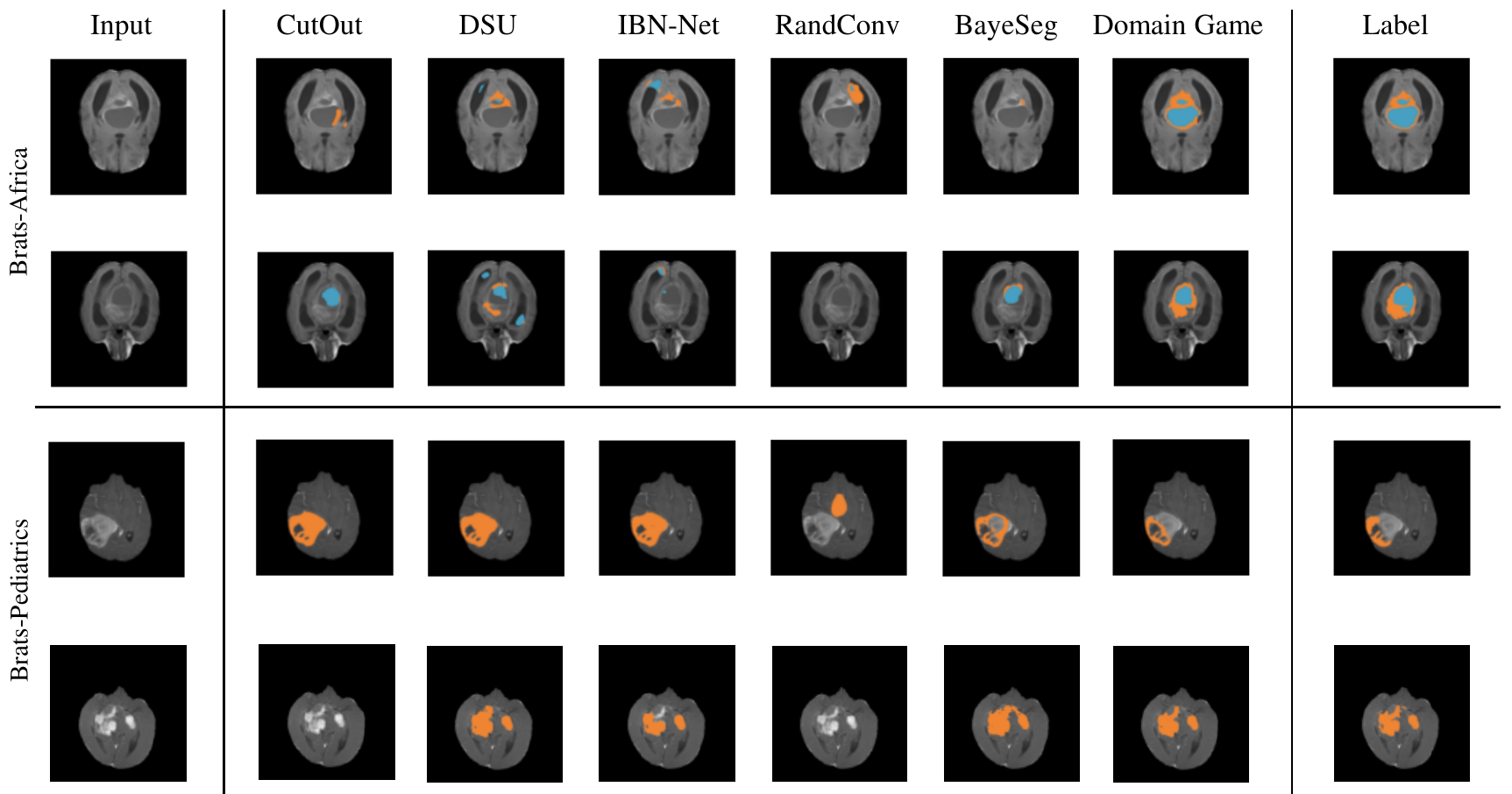}
    \caption{ \small Qualitative visualization of brain tumor segmentation.
    }
    \vspace{-0.6cm}
    \label{fig:brain}
\end{figure}


We showcase some qualitative visualizations of brain tumor segmentation results in Fig.\ref{fig:brain}. 
The design of the brain tumor experiments is guided by two primary objectives. Firstly, we seek to enable the model to generalize across different geographical regions, especially to areas with constrained medical resources (\ie the scarcity of specialized healthcare professionals and advanced medical infrastructure). 
We aim to achieve a model with robust domain generalization capabilities, which, when extensively trained in high-income settings, helps address these issues in regions with limited resources \cite{jha2023upending}.

Secondly, we aim for the model to generalize across diverse age groups and tumor subtypes. The development of deep learning models typically requires a large volume of high-quality data. However, pediatric gliomas (pediatric diffuse midline gliomas) are rarer, smaller, and more challenging to detect compared to adult cases (adult glioblastomas), resulting in poorer prognoses and higher mortality \cite{mackay2017integrated}. Thus, effectively generalizing models trained on adult data to pediatric cases could significantly enhance early tumor detection and assessment.


\vspace{1mm}
\noindent\textbf{Analysis:} Table \ref{tab:brain} presents the Dice and Jaccard scores. In summary, Domain Game achieves the highest average scores in both metrics on target domains, at \textit{65.05} and \textit{59.53}, respectively. This signifies an approximate \textasciitilde\textit{10.5}\% boost in Dice compared to the second-best score achieved by IBN-Net \cite{ibn-net},  and around \textasciitilde\textit{8.6}\% increase in Jaccard score surpassing the second-best BayeSeg \cite{BayeSeg}.



\vspace{-0.3cm}
\subsection{Module Ablation Study}
We perform ablation studies to investigate the contribution of modules in the Domain Game, and the results are summarized in Table \ref{tab:ablation}. We analyze each module by excluding it from the benchmark (last row). Eliminating the domain encoder (first row) shows a performance drop of more than  \textasciitilde\textit{18.6}\%. Removing the space constraint (second row) results in approximately a \textasciitilde\textit{6.0}\% decrease in performance. Finally, we test the usefulness of the geometry transformation strategy. The results show that removing the rotation transform (third row) and flip transform (fourth row) from the geometry transformation set cause \textasciitilde\textit{5.3}\% and \textasciitilde\textit{1.1}\% performance drop, respectively.

\begin{table*}[!tbp]
\caption{Analysis of component-specific ablation within the proposed method. The red number indicates declined performance compared with benchmark.
} %
\label{tab:ablation}

 \centering
	\resizebox{\linewidth}{!}{

    \begin{tabular}{l@{\hspace{0.3cm}}|cc|cc|cc}
     \toprule
    \multirow{2}{*}{\makecell[c]{Module Ablation\ \ }}
    &\multicolumn{2}{c|}{{\makecell[c]{Avg. on Prostate}}}
    &\multicolumn{2}{c|}{{\makecell[c]{Avg. on Glioma}}}
     &\multicolumn{2}{c}{{\makecell[c]{ \ \ Avg. of Both Tasks}}}\\

        & Dice & Jaccard & Dice & Jaccard  & Dice & Jaccard  \\
         \midrule 
     w/o domain encoder $\En$
       &69.72$\pm$09.1 \down{12.51}& 65.38$\pm$08.4 \down{14.81}
       & 51.19$\pm$25.7 \down{13.86} &  47.25$\pm$22.9 \down{12.28}
       & 60.46   \down{13.18} & 56.32 \down{13.54}\\
      w/o  space constraint
       &\ \ \  77.32$\pm$09.1 \down{05.31} \ \ \ &\ \ \  74.27$\pm$08.7 \down{05.92} \  \ \ 
       &\ \ \  60.89$\pm$23.9 \down{04.16} \ \ \ &  57.38$\pm$21.2 \down{02.15}
       & \ \  69.11  \down{04.53}  \ \ & 65.83   \down{04.03}
         \\

        w/o  rotation transformations
       &\ \  78.31$\pm$07.6 \down{03.92} \ \ &\ \  75.95$\pm$07.0 \down{04.68} \ \ 
       &\ \  60.37$\pm$24.7 \down{04.68} \ \ &  56.98$\pm$21.5 \down{02.55}
       & \ \  69.34   \down{04.30}  \ \ & 66.47  \down{03.39}  
         \\
        
        w/o  flip transformations
       &\ \  81.17$\pm$07.4  \down{01.06} \ \ &\ \  79.44$\pm$06.3 \down{00.75} \ \ 
       &\ \  64.22$\pm$23.7 \down{00.83} \ \ &  58.87$\pm$20.7 \down{00.66}
       &  \ \ 72.70   \down{00.94}  \ \ & 69.16  \down{00.70}  
         \\

        \midrule
        proposed  method (Benchmark)
       & \ \  \hspace{-0.1cm}\textBF{82.23$\pm$06.7}  \ \ &  \ \  \hspace{-0.1cm}\textBF{80.19$\pm$05.9 } \ \ 
       & \ \  \hspace{-0.1cm}\textBF{65.05$\pm$23.4 } \ \ &  \ \  \hspace{-0.1cm}\textBF{59.53$\pm$19.6 } \ \ 
       & \ \ \hspace{-0.1cm}\textBF{ 73.64  } \ \ &\hspace{-0.1cm}\textBF{ 69.86    }
         \\

     \bottomrule
    \end{tabular}
    }

\vspace{-0.5cm}
\end{table*}

\section{Conclusion}

We propose a new framework to tackle the domain shift when only one source domain is available, in which an efficient machanism is exquisitely designed to distangle diagnostic and domain-specific features, respectively. The new method is motivated by the observation of the different responses of these two distinct features to geometric transformations. Experimental results demonstrate the effectiveness and efficiency of the proposed method in improving the generalizability of a trained model to wider domains. We expect the proposed method can serve as a useful tool to alleviate the ubiquitous domain shift problem in real-world clinical applications.







%
%
%


\bibliographystyle{splncs04}
\bibliography{main.bib}

\newpage

\end{document}